\documentclass[letterpaper]{article} % DO NOT CHANGE THIS
\usepackage{aaai24}  % DO NOT CHANGE THIS
\usepackage{times}  % DO NOT CHANGE THIS
\usepackage{helvet}  % DO NOT CHANGE THIS
\usepackage{courier}  % DO NOT CHANGE THIS
\usepackage[hyphens]{url}  % DO NOT CHANGE THIS
\usepackage{graphicx} % DO NOT CHANGE THIS
\usepackage{natbib}  % DO NOT CHANGE THIS AND DO NOT ADD ANY OPTIONS TO IT
\usepackage{caption} % DO NOT CHANGE THIS AND DO NOT ADD ANY OPTIONS TO IT
\usepackage{algorithm}
\usepackage{algorithmic}
\usepackage{booktabs} % For formal tables
\usepackage{amsmath}
\usepackage{multirow}
\usepackage{bbding}
\usepackage{amssymb}
\usepackage{makecell}
\usepackage{bbding}

\usepackage{newfloat}
\usepackage{listings}
\DeclareCaptionStyle{ruled}{labelfont=normalfont,labelsep=colon,strut=off} % DO NOT CHANGE THIS
\floatstyle{ruled}
\newfloat{listing}{tb}{lst}{}
\floatname{listing}{Listing}
\title{Instruct-NeuralTalker: Editing Audio-Driven Talking Radiance Fields with Instructions}
\author{
    Yuqi Sun,
    Ruian He,
    Weimin Tan,
    Bo Yan
}
\affiliations{
 	School of Computer Science, Shanghai Key Laboratory of Intelligent Information Processing,\\
	Shanghai Collaborative Innovation Center of Intelligent Visual Computing, Fudan University, Shanghai, China
	
}

\usepackage{bibentry}
\begin{document}

\maketitle

\begin{abstract}
	Recent neural talking radiance field methods have shown great success in photorealistic audio-driven talking face synthesis. In this paper, we propose the first novel interactive framework that utilizes human instructions to edit talking radiance fields to achieve personalized talking face generation. Given a short speech video, we first build an efficient talking radiance field and then apply the latest conditional diffusion model to edit images with human instructions during the optimization. To ensure audio-lip synchronization during the editing process, we propose a progressive dataset updating strategy to ensure successful editing while preventing distortion of lip shapes. We also introduce a lightweight refinement network for complementing image details and achieving controllable detail generation in the final rendered image. Our method enables real-time rendering at up to 30FPS on consumer hardware. Multiple metrics and user verification show that our approach significantly improves rendering quality compared to state-of-the-art methods.
\end{abstract}

\section{Introduction}

%早期的方法利用一些中间人脸表示，如显式3D人脸模型，表情回归系数，或者2D landmarks来实现真实且可控的人脸生成。但是这些离散的中间表示会导致测量误差，从而引起原始语音信号和唇部运动之间的不匹配。

% 音频驱动的说话人脸生成是计算机视觉和图形学中一个长久存在的目标，它能够广泛应用在数字化人体，虚拟视频会议，VR/AR, 3D telepresence等领域。最近基于Neural Radiance Field(NeRF)的方法在实现真实且可控的说话人脸生成上取得了很大的进步，这些方法通过构建头部的隐式3D表示，将音频特征直接映射到动态神经辐射场来驱动脸部变形，从而生成生动连贯的说话人脸视频。尽管隐式3D人脸建模使得这些方法支持视角合成，背景替换，姿势控制等编辑任务，在这些方法上实现更加自由的编辑例如风格迁移是困难的而且未经尝试的。

%本文在这些方法的基础上，探索了一种更加广泛和自由的编辑方式：通过简单的人类指令编辑talking radiance filed来实现个性化的说话人脸生成。尽管如此， 
%在这些方法的基础上，本文提取了一种新颖的交互式编辑框架：editing talking radiance field with simple human instructions.
%Due to the implicit 3D face modeling, these method also enables editing tasks such as view synthesis and background replacement. Neveltheless, achieving more general editing target such as style transfer on these methods is challenging and unexplored.

%Given a speech video, our method first builds an original audio-driven talking radiance field and edits it with human instructions. \textit{Original Portrait Synthesis} shows the generated results of the original radiance field from the input audio. By giving editing instructions like \textit{"Make him look like an old man"} and \textit{"Turn him into Edward Munch painting"}, our methods can synthesize high-quality talking faces that meet the editing target well while maintaining audio-lip synchronization. The complete rendering pipeline can be done in real-time on consumer hardware. Benefiting from the 3D neural representation modeling, our method can easily perform other editing tasks like novel view synthesis and background replacement.

\begin{figure}[t]
	\centering
	\includegraphics[width=\linewidth]{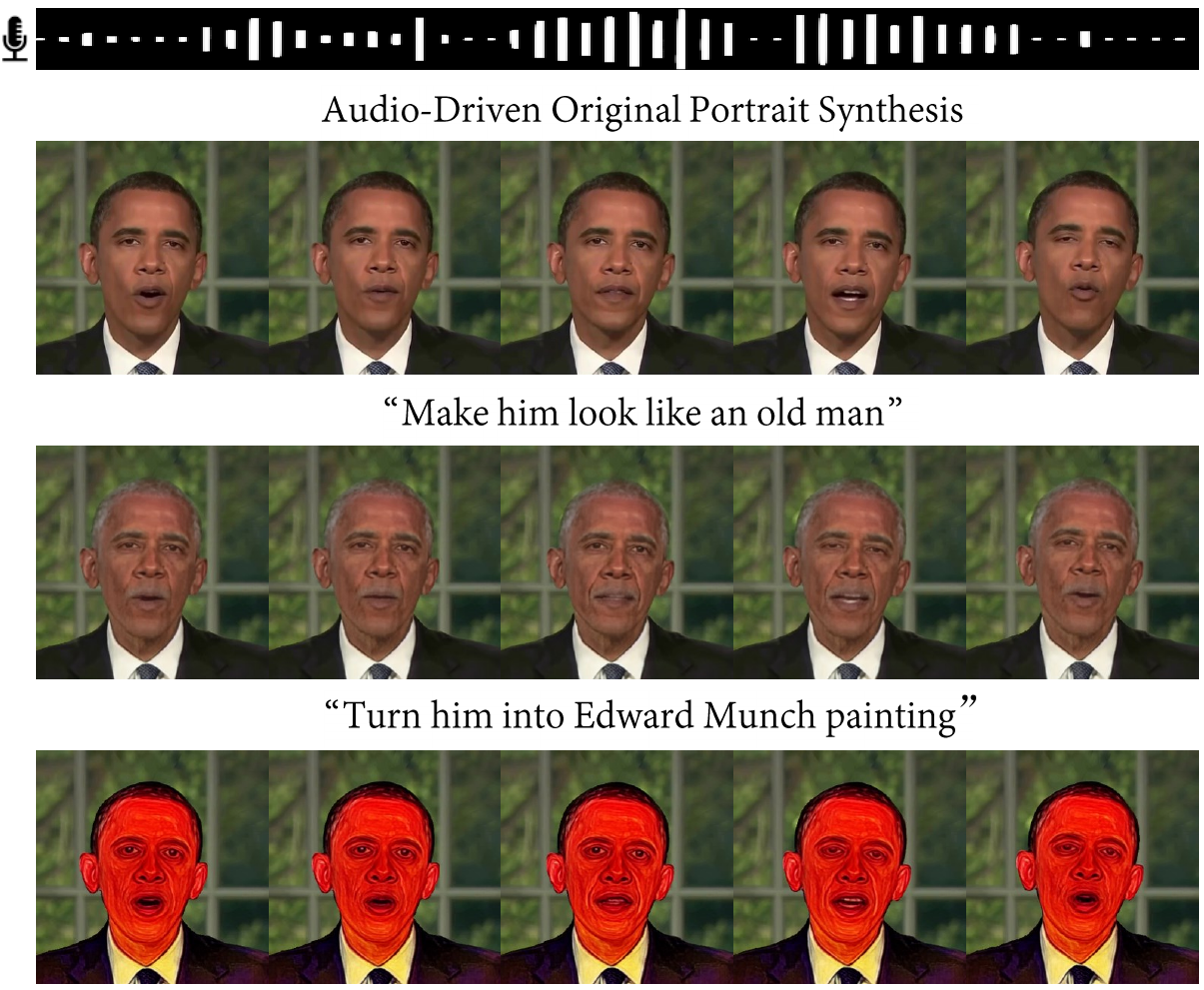}
	\caption{\textit{Audio-Driven Original Portrait Synthesis} shows the generated results of the original radiance field driven by the input audio. By giving editing instructions like \textit{"Make him look like an old man"} and \textit{"Turn him into Edward Munch painting"}, our methods can synthesize high-quality talking faces that meet the editing target well while maintaining audio-lip synchronization.}
	\label{intro-1}
\end{figure}

%\begin{figure*}[t]
%	\centering
%	\includegraphics[width=0.9\linewidth]{imgs/intro-2-crop.pdf}
%	\caption{\textbf{Illustration of controllable detail generation.} Given an instruction \textit{"Make him look like Van Gogh"}, our method generates talking faces with the style of Van Gogh. By manipulating the added weight of the refinement network's output, we can generate controllable detail, where lower weight values yield smooth results, and higher weight produces rich details.}
%	\label{detail generation}
%\end{figure*}

%\begin{figure*}[t]
%	\centering
%	\includegraphics[width=\linewidth]{imgs/intro-2-crop.pdf}
%	\caption{\textbf{Illustration of controllable detail generation.} When we input instruction \textit{"Make him look like Van Gogh"}, our method generates talking faces with the style of Van Gogh. By manipulating the added weight of the refinement network's output, we can generate controllable detail, where lower weight values yield smooth results, and higher weight produces rich details.}
%	\label{detail generation}
%\end{figure*}

Audio-driven talking face generation has been a long-standing task in computer vision and graphics, with widespread applications in digital humans, VR/AR, 3D telepresence, and virtual video conferencing. Recent methods \cite{Guo2021ADNeRFAD, Tang2022RealtimeNR}  have made significant progress in achieving photorealistic and controllable talking face generation based on Neural Radiance Fields (NeRFs) \cite{mildenhall2020nerf}. By constructing an implicit 3D face model, their approaches allow impressive editing capabilities such as novel view synthesis and background replacement. However, achieving advanced editing tasks like semantic manipulation and style transfer on implicit representations remains challenging. 

%However, achieving advanced editing tasks like semantic manipulation and style transfer on implicit representations still remain challenging. 

% 一些方法将图像风格化的方法引入NeRF编辑中, 他们通常使用一张参考图作为编辑目标提示，风格化3D内容的整体外观。紧接着，一些工作将图像提示拓展到文本，使用预训练的视觉语言模型，例如CLIP。ClipNeRF, NeRF-Art通过鼓励NeRF渲染结果和输入文本在CLIP空间的相似性来实现文本控制的编辑。非常最近，生成式模型如Stable Diffusion在text-to-image generation and editing领域取得了巨大的成功。受到他们成功的鼓舞，instruct nerf2nerf实现了令人印象深刻的基于人类指令的NeRFs编辑，使用l instructPix2Pix, 一个基于指令进行图像编辑的diffusion model.

Some methods \cite{Huang2022StylizedNeRFC3, NguyenPhuoc2022SNeRFSN, Zhang2022ARFAR} introduce image stylization methods into NeRFs editing. They usually take a reference image as condition to stylize the global scene appearance of the 3D content. ClipNeRF \cite{Wang2021CLIPNeRFTD} and NeRF-Art \cite{Wang2022NeRFArtTN} extend image conditions to text prompt by leveraging pre-trained visual language models like CLIP \cite{Radford2021LearningTV}. More recently, Instruct-NeRF2NeRF (in2n) \cite{Haque2023InstructNeRF2NeRFE3} achieves impressive instruction-based editing by incorporating a diffusion model InstructPix2Pix (ip2p) \cite{Brooks2022InstructPix2PixLT}. Nevertheless, all these methods focus on static scenes and have not been fully explored in dynamic scenarios. In this work, we propose the first interactive framework, Instruct-NeuralTalker, for semantically editing dynamic Talking Radiance Fields (TRFs) with human instructions, which enables highly personalized talking face generation.

\begin{figure*}[t]
	\centering
	\includegraphics[width=0.95\linewidth]{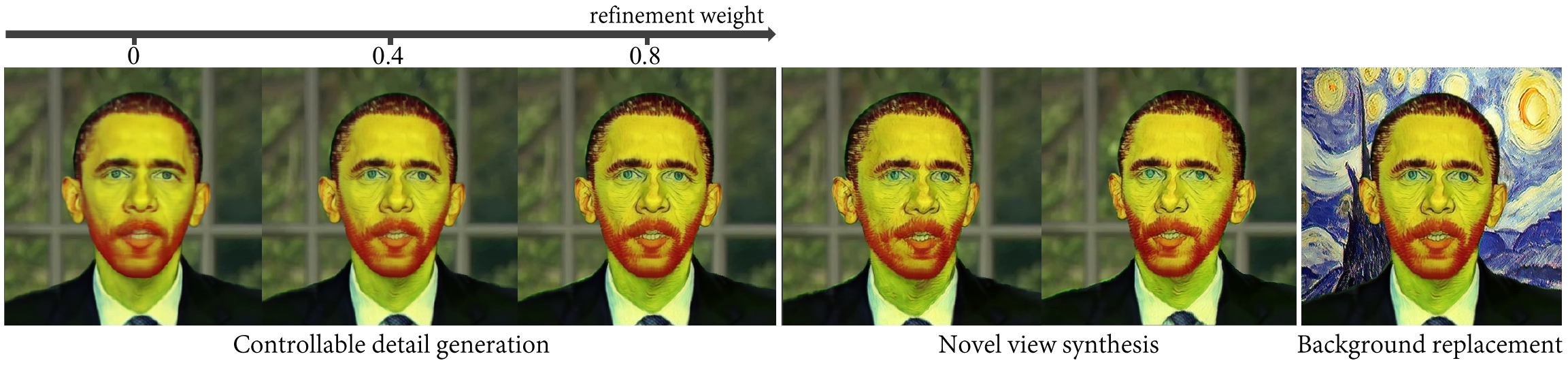}
	\caption{Visual results for instruction \textit{"Make him look like Van Gogh"}. By manipulating the added weight of the refinement network's output, we can generate controllable details, where lower weight values yield smooth results, and higher weight produces rich details. Our method also supports other editing tasks like novel view synthesis and background replacement.}
	\label{detail generation}
\end{figure*}

Given a short speech video, our approach first builds an initial TRF for the original sequence. Then users can provide text instructions specifying personalized editing goals, such as \textit{"Make him look like an old man"} or \textit{"Turn him into Edward Munch painting"}. Once the desired appearance is defined, we progressively modify the training dataset using ip2p during the optimization to guide the initial TRF toward the edited target. Ultimately, we obtain a TRF with an "old man" or "Edward Munch painting" style that can generate corresponding talking faces driven by audio. Leveraging the latest efficient NeRF architecture, our method achieves real-time rendering on consumer hardware up to 30FPS. We showcase some visual results in Figure \ref{intro-1}.

In contrast to editing static scenes, it is essential to consider motion consistency in editing dynamic radiance fields. For talking faces generation, it involves ensuring audio-lip synchronization in the edited results. 
Applying ip2p to update datasets like in2n may lead to noticeable changes in lip shape, which we attribute to the large text guidance scale and steps of the reverse diffusion process in ip2p. To address this, we propose a progressive dataset updating strategy that gradually increases the text guidance scale and diffusion steps during the optimization. This approach helps the model preserve lip shape in a coarse-to-fine manner. We also estimate the lip parsing mask and calculate a lip-edge loss to restrain the edges of the lips for audio-lip synchronization.

On the other hand, since the editing results of ip2p do not guarantee cross-view consistency, the edited TRF tends to produce over-smoothed results. To solve this, we introduce a refinement network for complementing image details. We weighted add the rendered result from the edited TRF and the result of the refinement network to produce the final output. It also enables us to generate controllable detail by adjusting the refinement weight. Our method supports 3D editing tasks like novel view synthesis and background replacement. Figure \ref{detail generation} illustrates the results. We also implement an interactive interface facilitating users to perform instruction editing. We highly recommend watching the video in the supplementary material for a better experience. 

%Figure \ref{intro-2} shows our results with different refine weights. When the weigth is set as 0, the refinemenet network is disabled and the output from talking radiance field missing image details 

%In this paper, we explore a more general and freeform editing method by using simple human instrcution to edit the talking radiance field for personalized talking face generation, building upon the existing methods.

%In summary, this paper mainly has the following contributions:
In summary, this paper has the following contributions:

% itemize
%\begin{itemize}
%\item We propose Instruct-NeuralTalker,  the first interactive framework that edits the audio-driven talking radiance field with simple human instructions to achieve personalized talking face generation, which can support general editing tasks including instruct-editng, novel view synthesis and background replacement. In addition, Instruct-NeuralTalker enables real-time rendering on consumer hardware.
%
%\item In order to ensure audio-lip synchronization, we develop an iterative dataset updating strategy and a lip-edge loss to constrain the lips change of the editing results during the optimization process.
%
%\item We introduced a lightweight refinement network to complement  high-frequency image details and achieve controllable detail generation in the rendering process.
%
%\item Extensive experiments demonstrate the superiority of our method in video quality compared to the state-of-the-art.
%
%\end{itemize}

%\noindent
$\bullet$ We propose Instruct-NeuralTalker, the first interactive framework to semantically edit the audio-driven talking radiance fields with simple human instructions. It enables real-time rendering on consumer hardware.

%\noindent
$\bullet$ In order to ensure audio-lip synchronization, we develop a progressive dataset updating strategy to prevent distortion of lip shape during the optimization process.

%\noindent
$\bullet$ We introduce a lightweight refinement network to complement high-frequency image details and achieve controllable detail generation in the rendering process.

%\noindent
$\bullet$ Extensive experiments demonstrate the superiority of our method in video quality compared to the state-of-the-art.

% Head 1
\section{Related work}
\label{sec::rw}

\textit{\textbf{Audio-Driven Taking Face Generation.}} Audio-driven talking face generation aims to reenact a specific person's speaking video based on audio input. Traditional methods introduce an explicit 3D representation to achieve controllable face generation, such as 3D morphable face models \cite{Thies_Elgharib_Tewari_Theobalt_Nießner_2020} and facial landmarks \cite{Zakharov_Shysheya_Burkov_Lempitsky_2019}. In addition, some methods \cite{10.1145/3394171.3413532, Zhou2020MakeItTalkST, zhang2022sadtalker} focus on generating talking faces from a single image. However, the explicit representation may lead to error accumulation and mismatches between audio and lips. With the development of NeRF, recent methods propose to model talking faces as a dynamic talking radiance field \cite{Pumarola_Corona_Pons-Moll_Moreno-Noguer_2021} for reducing information loss. AD-NeRF \cite{Guo2021ADNeRFAD} was the first to directly map audio features to dynamic NeRFs for lips movement generation, achieving better audio-visual consistency. DFRF \cite{Shen2022LearningDF} can adapt to new identities with limited reference images through pretraining on a large dataset. DFA-NeRF\cite{Yao2022DFANeRFPT} decouples lip movement features related to audio from personalized attributes unrelated to audio using a self-supervised learning approach. RAD-NeRF \cite{Tang2022RealtimeNR} leverages recent successful grid-based NeRF approaches \cite{Müller_Evans_Schied_Keller_2022} to model dynamic head movements using efficient audio and spatial grids, which achieves real-time talking face generation.

\begin{figure*}[t]
	\centering
	\includegraphics[width=0.9\linewidth]{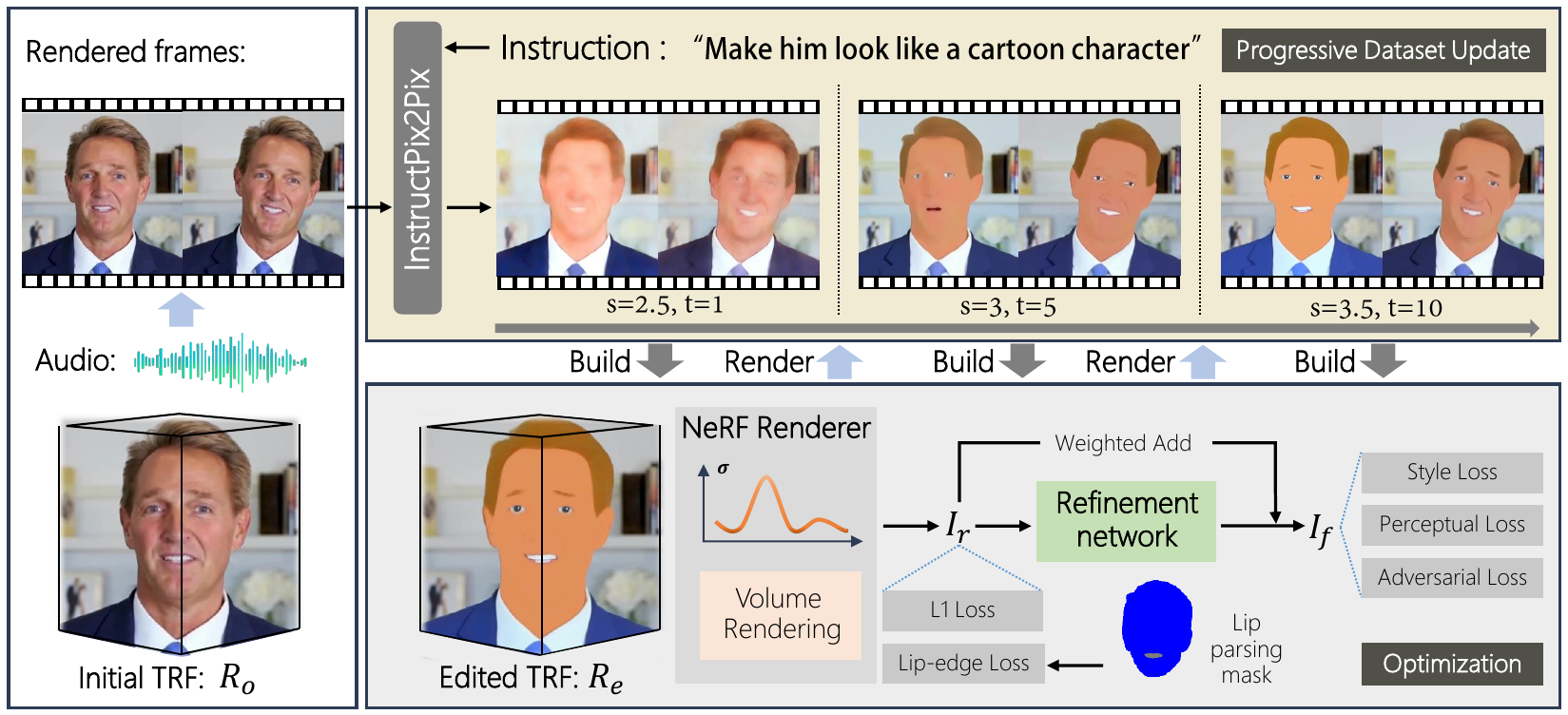}
	\caption{\textbf{Overview of our pipeline.}  Given a text instruction such as "Make him look like a cartoon character," we use InstructPix2Pix (ip2p) to edit the rendered frames from initial TRF $\mathbf{R_o}$ to get cartoon-style images, which are used to build the edited TRF $\mathbf{R_e}$. Our progressive dataset update strategy keeps increasing the text guidance $\textbf{s}$ and diffusion steps $\textbf{t}$ in ip2p to control the degree of editing. A refinement network is proposed for compensating high-frequency image details. The final image $I_f$ is obtained by weighted-add the output of our refinement network and rendered image $I_r$ from $\mathbf{R_e}$.}
	\label{method}
\end{figure*}

\textit{\textbf{Instruction-based Neural Editing.}}
The rise of large-scale language models LLMs (such as ChatGPT) has enabled complex task control through natural language, particularly instructions\cite{Ouyang2022TrainingLM}. InstructPix2Pix \cite{Brooks2022InstructPix2PixLT} demonstrates remarkable results in instruction-based editing for 2D image tasks, leveraging the GPT \cite{Brown2020LanguageMA} and stable diffusion \cite{rombach2021highresolution}. Building upon this progress, Instruct-Nerf2NeRF (in2n) \cite{Haque2023InstructNeRF2NeRFE3} introduces ip2p to 3D scene editing and allows for intuitive editing for the first time, showcasing the potential of instruction-based control in NeRF editing. Nevertheless, they focus on editing static scenes. Inspired by in2n, our method first introduces instruction-based editing into dynamic NeRFs to achieve a personalized generation of talking faces. It gives rise to an interface entirely controlled by language, facilitating a more comprehensive range of intuitive 3D talking face editing. 

%AvatarCLIP\cite{Hong2022AvatarCLIPZT} employs a shape VAE network to generate coarse mesh shapes and motions directly from text and use CLIP for complex texture control and generation. 
%
%Recent works have introduced pretrained language-visual models into the training process of NeRF to achieve text-based NeRF editing. ClipNeRF \cite{Wang2021CLIPNeRFTD} and NeRF-Art \cite{Wang2022NeRFArtTN} encourage similarity between the scene and text embeddings from CLIP \cite{Radford2021LearningTV} to enable text-based control. AvatarCLIP\cite{Hong2022AvatarCLIPZT} employs a shape VAE network to generate coarse mesh shapes and motions directly from text and use CLIP for complex texture control and generation. 

%Distilled Feature Fields\cite{Kobayashi2022DecomposingNF} and Neural Feature Fusion Fields \cite{Tschernezki2022NeuralFF} distill 2D features from pretrained teacher networks such as LSeg\cite{Li2022LanguagedrivenSS} and DINO \cite{Caron2021EmergingPI} and inject them into 3D feature fields that are jointly optimized with NeRF.

\section{Method}
\label{sec:method}

% 我们的方法使用一个短的约4分钟的说话视频作为输入，并通过人脸解析和人脸追踪获得人脸解析mask、头部位姿等。
% 给定一个初始的Talking Radiance Field和一个编辑指令，例如“Make him look like a cartoon character”, 我们的目标是得到一个编辑后的
%talking radiance field，能够通过音频驱动渲染出符合编辑指令的说话人脸。我们提出了Instruct-NeuralTalker, 一个新颖的互动框架来编辑dynamic talking radience field同时保持音唇一致性. 

%Given an initial Talking Radiance Field (TRF) and an editing instruction such as "Make him look like a cartoon character", we aim to obtain an edited TRF that can render taking faces in accordance with the given editing instruction, driven by audio input.

% 我们提出了Instruct-NeuralTalker, 一个新颖的互动框架来编辑dynamic talking radience field同时保持音唇一致性. 我们的方法使用一个短的约4分钟的说话视频作为输入，并通过人脸解析和人脸追踪获得人脸解析mask、头部位姿等。我们首先遵从最近的隐式说话人脸建模方法构建一个初始的Talking radiance field. 然后给定一个编辑指令，例如“Make him look like a cartoon character”, 我们利用ip2p迭代式编辑渲染的结果通过音频驱动，并将其作为训练数据优化talking radiance field. 最终，我们可以得到一个编辑后的神经辐射场，能够通过音频驱动生成符合编辑指令的说话人脸。

We present Instruct-NeuralTalker, a novel interactive framework for editing dynamic talking radiance fields (TRFs). Our approach takes as input a short talking video along with its corresponding source data, such as face-parsing masks and head poses obtained through face parsing and tracking. Following recent implicit talking face modeling methods, we first build an initial TRF $\mathbf{R_o}$. Additionally, given an editing instruction, e.g., \textit{"Make him look like a cartoon character"}, we use ip2p to iteratively edit the audio-driven rendered result from $\mathbf{R_o}$ and use it as training data in turn to optimize the talking radiance field. In the end, we can obtain an edited TRF  $\mathbf{R_e}$ that can generate audio-lip-synchronized talking faces that conform to the editing instructions.

\subsection{Preliminaries}
\label{sec:method:pre}
% 我们的方法建立在几个最新的研究进展之上。首先，我们使用RAD-NeRF作为建立说话人脸radiance field的骨干网络，它使用了高效的NeRF结构支持实现快速训练和实时渲染。其次我们从in2n中得到启发，使用图像编辑中的扩散模型的最新进展ip2p来编辑训练集图像，通过在优化过程中进行数据更新来编辑talking radiance field. 我们在下面简要介绍这些方法，并说明我们方法的不同之处。

%Our method builds upon several recent research advancements. Firstly, we employ RAD-NeRF as the backbone network for constructing the radiance field of talking faces. Secondly, inspired by in2n, we leverage the latest advancements in conditional diffusion models ip2p for dataset editing and update the talking radiance field during optimization. In the following, we provide a brief overview of these methods and highlight the distinctive aspects of our approach.

% RAD-NeRF通过引入efficiency NeRF领域中的最新进展grid-based nerf架构，到talking radiance field中来实现实时渲染。Grid-based NeRF将 3D 场景特征存储在显式 3D特征网格结构，并使用高效的数据结构如multi-resolution hash table和low-rank tensor components来实现更快速的查询和线性插值。为了将音频特征引入3D特征网格同时避免高维特征的the curse of dimensionality，RAD-NeRF将音频和头部分别建模。具体来说，给定空间中一点x
\textit{RAD-NeRF.}
We employ RAD-NeRF\cite{Tang2022RealtimeNR} as the backbone network for building TRFs. It directly maps audio features to dynamic neural radiance fields representing the talking face. RAD-NeRF achieves real-time rendering by incorporating the latest advancements in the grid-based NeRF architecture, which stores explicit 3D features in efficient data structures such as multi-resolution hash tables \cite{Müller_Evans_Schied_Keller_2022} and low-rank tensor components \cite{Chen2022TensoRFTR} for efficient computation. To be specific, RAD-NeRF encodes a 3D coordinate into spatial feature $\textbf{f}$ with a 3D feature grid encoder along with a spatial-dependent audio vector $\textbf{a}$, an eye feature $e$ and appearance embedding $i$. The plenoptic function is modeled by $ \mathcal{F} : \textbf{c}, \sigma = \Theta(\textbf{f}, \textbf{a}, e, i) $, where $\textbf{c}$ is the emitted color, $\sigma$ is the volume density and $\Theta$ is the weight of a small MLP. The final color of the pixel is obtained by volume rendering:

\begin{equation}
	\textbf{C}(\textbf{r}) = \displaystyle \sum_{i}T_i \alpha_i \textbf{c}_i, T_i = \prod_{j<i}(1 - \alpha_j)
	\label{volume rendering}
\end{equation}
where $\alpha_i = 1 - exp(-\sigma_i \delta_i)$ is the opacity, $T_i$ is the transmittance, and $\delta_i = t_{i+1} - t_i$ is the step size between two 3D coordiantes.

% 去噪扩散模型是一种生成式方法，通过正向扩散过程来逐渐向数据添加随机噪声，然后通过学习反向扩散过程，从噪声中构建所需的数据样本。InstructPix2Pix (ip2p)的目标是使用diffusion model实现基于指令的图像编辑。给定一张图片，一个文本编辑指令，和一个纯噪声图像，模型使用一个U-Net在每次step中估计前向过程中添加的随机噪声:
% 通过控制Classifier-free Guidance
%The denoising diffusion model is a generative method that gradually adds random noise to the data through a forward diffusion process, and then builds the desired data distribution from the noise by learning a backward diffusion process. 
%Denoising diffusion model

\textit{InstructPix2Pix.}
InstructPix2Pix (ip2p) aims to achieve instruction-based image editing using the diffusion model, which learns to gradually denoise pure Gaussian noise into data samples. Given an original image x, a text instruction $c_T$, and an image conditioning $c_I$, ip2p learn a network $\epsilon_\theta$ that predicts the noise added to the noisy latent $z_t$. The network is optimized by the following loss function:

\begin{equation}
	L = \mathbb{E}_{\xi(x), \xi(c_I), c_T, \epsilon \sim \mathcal{N}(0, 1), t} [ \lvert\lvert \epsilon - \epsilon_\theta(z_t, t, \xi(c_I), c_T) \rvert\rvert^2_2 ]
	\label{diffusion}
\end{equation}

where $t \in (0,T)$  means denoising level. Since there exist two conditions $c_T$ and $c_I$, ip2p use two classifier-free guidance scales \cite{Ho2022ClassifierFreeDG}, $s_T$ and $s_I$, which can be adjusted to trade off how strongly the generated samples correspond with the edit instruction and how strongly they correspond with the input image.

\begin{figure*}[t]
	\centering
	\includegraphics[width=0.9\linewidth]{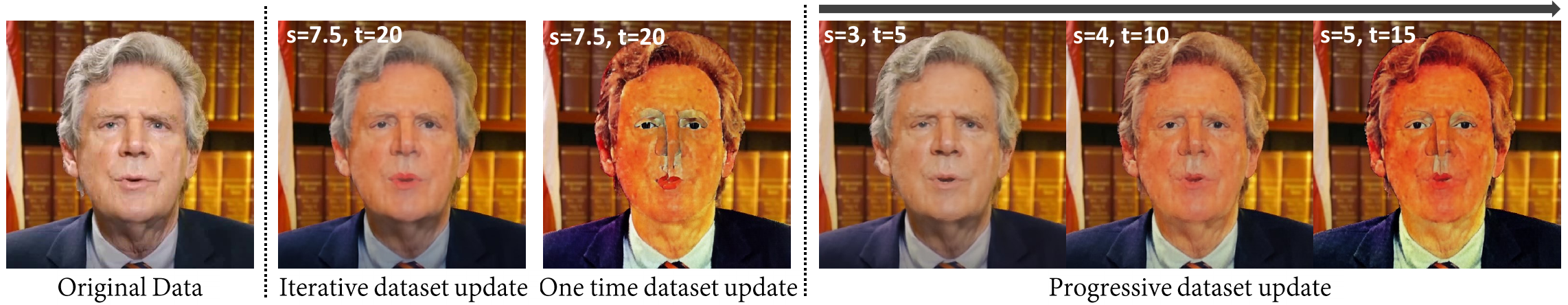}
	\caption{Results of different dataset update strategies with instruction "Turn him into a Modigliani painting". We show the results after 45k iterations. "Iterative dataset update" leads to little difference with original data, and "One time dataset update" destroys lip shape. Our method performs effective editing in a coarse-to-fine manner while maintaining the correct lip shape.}
	\label{pdu}
\end{figure*}

% 这个显示了我们方法基于指令进行编辑的能力。给定一个简单的编辑指令，我们的方法能够将原始序列编辑成符合目标的高质量说话人脸，并且保持出色的音唇同步性。

%我们使用RAD-NeRF作为骨干网络来搭建一个初始角色的talking radiance field，然后使用

%Instruct-NeuralTaker的目标是基于指令来编辑talking radiance field从而允许除了视角合成和背景替换之外的更加广泛和自由的说话人脸编辑。图3展示了我们的pipeline. 给定一段audio和对应的video frame, 我们使用RAD-NeRF来构建某个人的talking radiance field, 并使用ip2p根据text instruction编辑训练集中的图像来编辑神经辐射场。为了尽量保持嘴型，我们引入了Progressive Dataset update strategy迭代式地更新训练集。由于神经辐射场渲染出的图像缺少纹理细节，我们引入一个refinement network补充图像细节，并实现可控的细节生成。最后我们还引入了一个lip-edge 损失进一步控制嘴型来保持音唇一致性。优化完成后，Instruct-NeutalTalker可以在消费级硬件上实现实时的渲染。
\subsection{Instruct-NeuralTalker}
Our method aims to edit the talking radiance field based on text instructions. Figure \ref{method} illustrates our pipeline. Given audio and an initial TRF $\mathbf{R_o}$, we rendered $N$ frames ${\{I_i\}^N_{i=1}}$ from TRF driven by input audio as a training dataset. We then use ip2p to edit each frame in the dataset to get ${\{I_i^e\}^N_{i=1}}$ based on text instructions.
Since ip2p may destroy the lip shape, we introduce a progressive dataset update strategy to update the training set iteratively. After that, we take the new dataset to build the edited TRF $\mathbf{R_e}$, where we obtain the rendered result through volume rendering. Since the results are over-smoothed, we introduce a refinement network to enhance image details and achieve controllable detail generation. In addition, we incorporate a lip-edge loss to control lip shape further and maintain audio-lip synchronization.

\subsection{Progressive Dataset Update}
\label{sec:method:pdu}
In2n proposes an iterative dataset updating strategy that takes the original image as an extra condition to slowly edit training data (one image at a time). Although it works well on static scenes, it often fails when editing talking faces such that the edited image makes little difference from the original image. We attribute it to the extra image conditions in the diffusion model and the sparse views in the talking face. It is useful when we remove the extra image condition and perform one time dataset update, but a fixed large text guidance scale $s_T$ and diffusion step $t$ may destroy lip shape, as shown in the left of Figure \ref{pdu}. To solve this problem, we propose a progressive dataset update (PDU) strategy that uses a small increasing text guidance scale and diffusion step to update the dataset several times. In this way, PDU achieves a coarse-to-fine editing effect, which helps maintain audio-lip synchronization, shown in the right of Figure \ref{pdu}.

%Specifically, in the beginning, we set a parameter group $\{(\mathbf{s_0},\mathbf{t_0}),...,(\mathbf{s_k}, \mathbf{t_k})\}$with the gradual, incremental text guidance scales and diffusion steps for updating the training dataset $k$ times. 
To be specific, we fix the image guidance scale $s_I$, and only change the text guidance scale $s_T$ (denote as $s$) and diffusion step $t$. At first, we set a lower and upper bound $[s_l, s_u]$, $[t_l, t_u]$ for $s$ and $t$, respectively. We then determine text guidance scale $s_i$ and diffusion step $t_i$ for each dataset update by uniformly sampling:

\begin{equation}
	s_i = s_l + \frac{i - 1}{K - 1} (s_u - s_l), t_i = t_l + \lfloor \frac{i - 1}{K - 1} (t_u - t_l) \rfloor
	\label{equ.sample}	
\end{equation}
where $\lfloor \cdot \rfloor $ means rounding down, $i \in [1, K]$ and $K$ denotes how many times to update the training dataset ${\{I_i\}^N_{i=1}}$. We initialize $s_l, t_l$ with a small value and edit each frame in training dataset to obtain an edited dataset ${\{I^e_i\}^N_{i=1}}$, which we use as supervision in the optimization for building the edited talking radiance field $\mathbf{R_e}$ later. After that, we stop the optimization and re-render $N$ frames ${\{I^{'}_{i=1}\}_i^N}$ from $\mathbf{R_e}$. We then send them into ip2p for the second update with $s_2$ and $t_2$. After $K$ times dataset updates and optimizations, we obtain the final talking radiance field that satisfies the instruction. Figure \ref{pdu} illustrates an example when $K = 3$.

\begin{figure*}[t]
	\centering
	\includegraphics[width=0.9\linewidth]{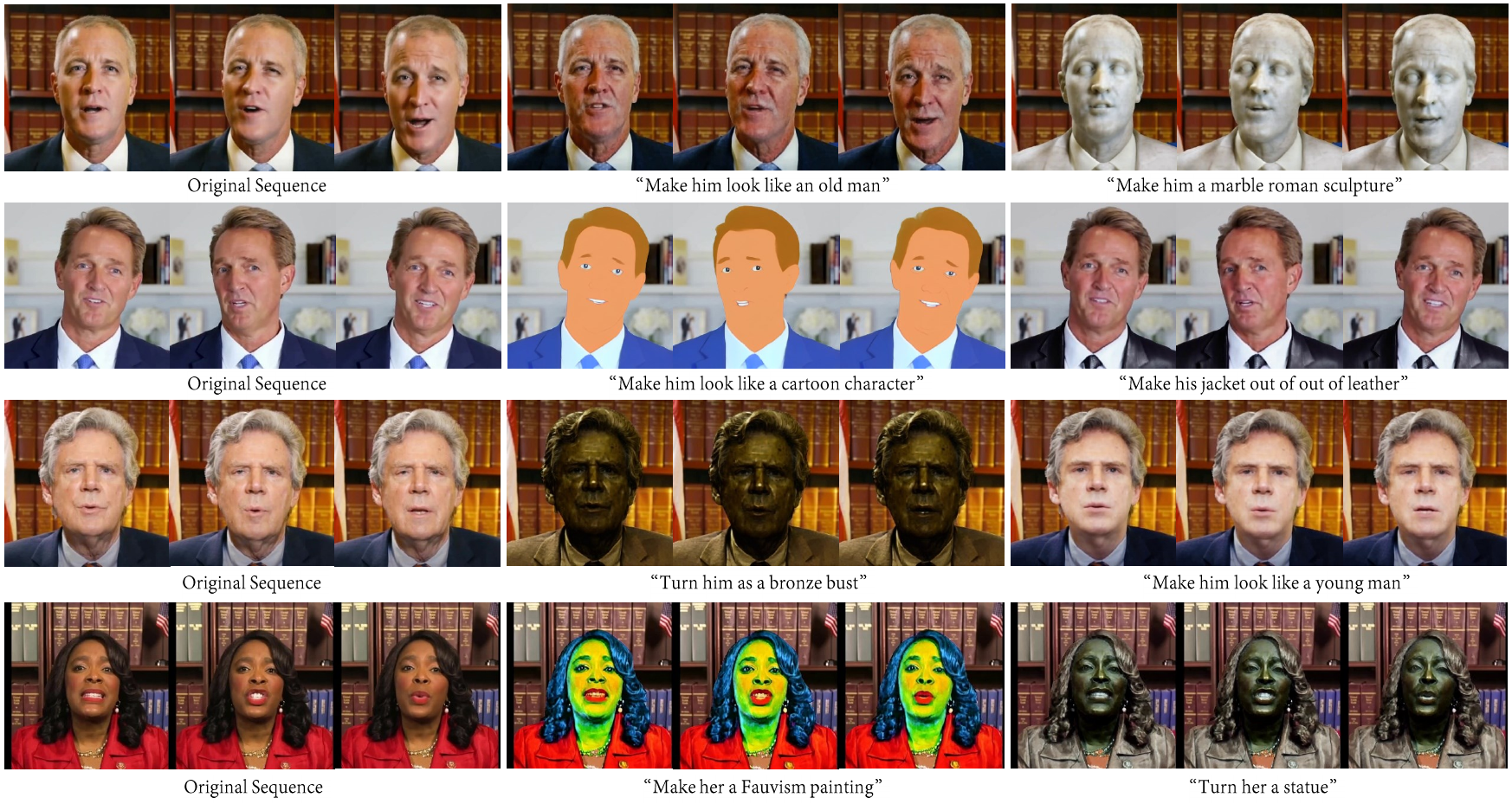}
	\caption{Qualitative results of our method. Given a simple text instruction, our method is able to edit the original sequence into a high-quality talking face that matches the editing target and maintains excellent audio-lip synchronization. }
	\label{result}
\end{figure*}

%
%\begin{equation}
%	\begin{aligned}
	%		& s = k * x + s_0, \\
	%		& t = m_1 * x^2 + m_2 * x + t_0
	%	\end{aligned}
%	\label{equ.3}	
%\end{equation}
%
\begin{figure*}[t]
	\centering
	\includegraphics[width=0.9\linewidth]{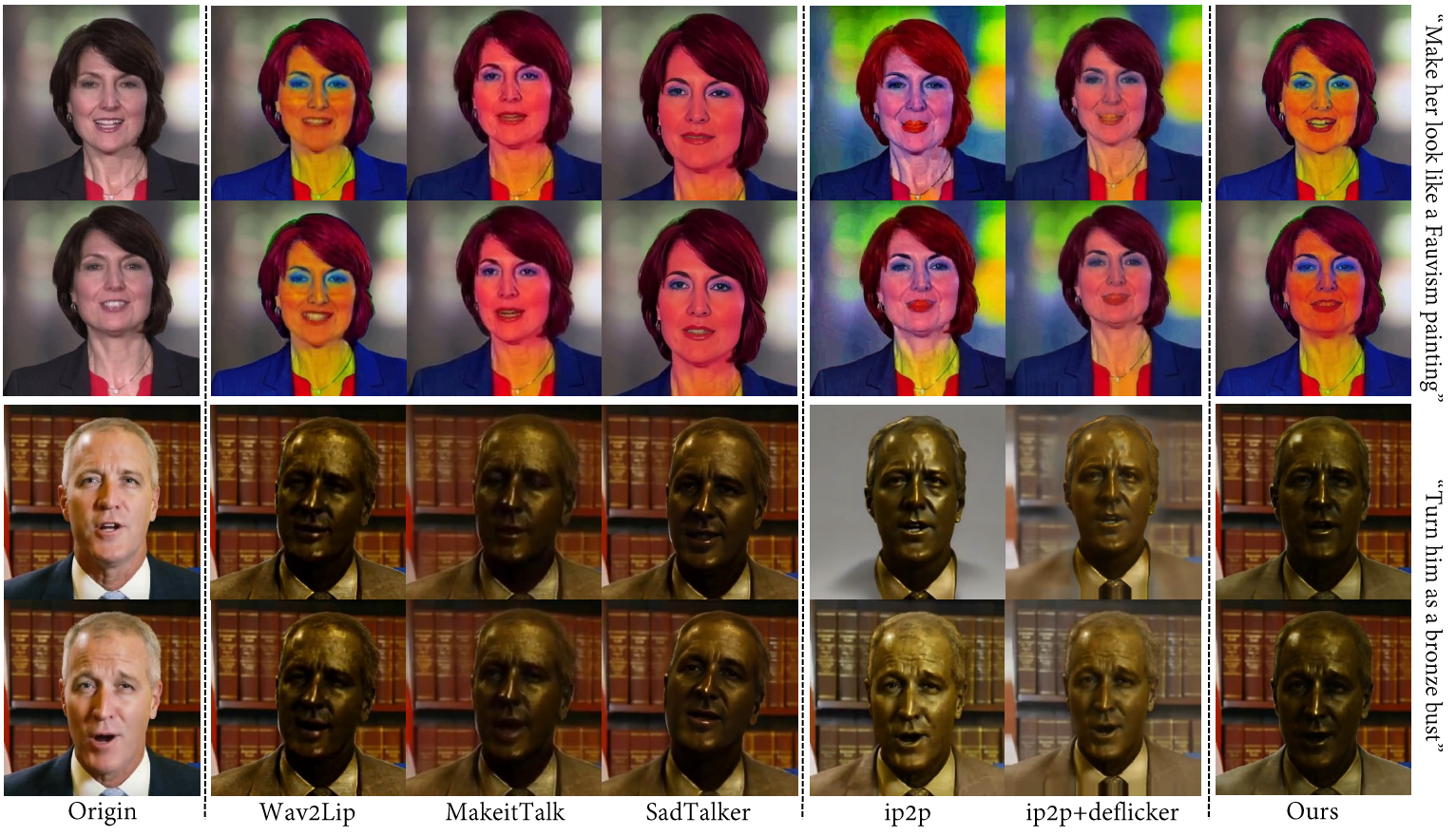}
	\caption{Qualitative comparisons with the state-of-the-art methods. Wav2Lip only generates static heads. The lip shapes of MakeitTalk and SadTalker are far from that of the origin sequence. The two post-processing based methods ip2p, ip2p+deflicker suffer from significant flicker and distortion. Our method was able to generate the best results.}
	\label{comparison}
\end{figure*}

%Although the edited talking radiance field is able to render satisfactory results with great temporal consistency, the images appear overly smooth and lack fine details. We attribute this to the fact that the network capacity of MLP limits its ability to learn high-frequency signals. Recently, convolutional-based deep learning methods have achieved remarkable performance in image enhancement tasks. Therefore, we introduce a refinement network $g$ that utilizes convolutional layers to complement details in the talking face images. Specifically, we draw inspiration from the Residual in Residual Dense Block (RRDB) in the super-resolution method ESRGAN \cite{Wang2018ESRGANES}, which consists of multi-level residual networks and dense skip connections. The multiple connections layers help the network capture the high-frequency details. To ensure a lightweight implementation, the refinement network utilizes only a single RRDB as the backbone network, resulting in a compact size of just 1MB. We send the result rendered from the talking radiance field $I_r$ into the refinement network and add the output to $I_r$ with the following equation:

% 虽然编辑之后的talking radiance field 已经能够渲染不错的满足时序一致性的编辑结果，图像过于平滑缺乏细节。我们认为将原因归结为MLP的网络容量限制了它学习高频信号的能力。最近基于卷积的深度学习方法在图像增强领域取得了出色的效果，因此我们引入了一个使用卷积的增强网络来补充说话人脸图像细节。具体来说，我们借鉴了超分方法ESRGAN中生成器中Residual in Residual dense block (RRDB)，由多个尺度的残差网络以及稠密连接组成。多层的连接有助于网络捕捉高频的细节，这正是我们需要的。为了保证轻量，refinement network只使用了一个RRDB。我们将refinement network的结果与talking radiance filed直接渲染的结果按权重相加得到最终的图像，公式如下：
\subsection{Refinement Network}
\label{sec:method:refine}
We render images $I_r$ from $\mathbf{R_e}$ by volume rendering, described in equation \ref{volume rendering}. Although the edited talking radiance field can render satisfactory results with great temporal consistency, the images appear overly smooth and lack fine details. We attribute this to the inconsistent cross-view editing results by ip2p and the network capacity of MLP, which limits its ability to learn high-frequency signals. Recently, some methods \cite{Chan2021EfficientG3} incorporate convolution networks in neural rendering to improve image quality. Inspired by this idea, we separate the rendering result into low-frequency and high-frequency signals and model them using NeRF $\mathtt{\mathcal{F(\cdot)}}$ and a convolution-based refinement network $\mathtt{g(\cdot)} $, respectively. The optimization objective can be written as:

\begin{equation}
	\mathop{\arg\min}\limits_{\mathtt{\mathcal{F}}, \mathtt{g}} = \mathcal{L}_{low}(\mathtt{V(\mathcal{F})} ) + \mathcal{L}_{high}( \mathtt{g(V(\mathcal{F}))})
	\label{equ.fre}	
\end{equation}
where $\mathtt{V}$ denotes the volume rendering equation, $ \mathcal{L}_{low} $ and $\mathcal{L}_{high}$ are the losses we modeled for low-frequency and high-frequency information.

Specifically, we implement a refinement network with a residual dense block consisting of multi-level residual networks and dense skip connections. The multiple connections layers help the network capture the high-frequency details. To ensure a lightweight implementation, the refinement network utilizes only a single block as the backbone network, resulting in a compact size of just 1MB. More details are represented in the supplementary material. We send the result rendered from the talking radiance field $I_r$ into the refinement network and add the output to $I_r$ with the following equation:
\begin{equation}
	I_f = \omega \mathtt{g}(I_r) + (1 - \omega) I_r
	\label{equ.3}	
\end{equation}
This allows us to achieve controllable detail generation by choosing different weights $\omega$. We show the result in Figure \ref{detail generation}. In addition, since the refinement network is lightweight, we can perform real-time rendering on consumer hardware.

% 我们希望它能够并生成与$I^e$相似的面部结构保持正确的唇部形状，因此我们使用了一个Lip-edge loss个和reconstuction loss. For $I_f$, 我们希望它能够从编辑图像学习纹理等高频细节，所以我们使用特征level上的损失，包括感知损失，风格损失和对抗损失。
\subsection{Losses}
\label{sec:method:losses}
Given an edited image $I^e$ as supervision, we apply different losses for the rendered result $I_r$ from the talking radiance field and the final result $I_f$. As described in the refinement network, we expect NeRF to model low-frequency information. Therefore, $\mathcal{L}_{low}$ consists of a reconstruction and lip-edge loss for $I_r$. For $I_f$, since we hope it can learn high-frequency details, we adapt $\mathcal{L}_{high}$ losses at the feature level, including perceptual loss, style loss, and adversarial loss.

\noindent
\textit{\textbf{Lip-Edge Loss.}} 
% 我们引入唇形边缘的约束进一步控制编辑后talking radiance field生成正确的嘴型。我们首先对原始人脸$I_i$进行解析，提取唇部的mask,然后在唇部的patch内部约束$I_r^{lip}$的边缘和$I_i^{lip}$中正确的边缘一致，公式如下：
We introduce constraints on the lip edges to further control the edited talking radiance field to generate the correct lip shape. For this sake, we borrow the depth smooth loss used in depth estimation \cite{Yin2018GeoNetUL} and apply it to the lip constraint. We first parse the original face to extract the lip mask $M^{lip}$, and then constrain the edges $I_r^{lip}$ inside the lip patch of $I_r$ to be consistent with the correct edges in $I^{lip}$, with the following equation:
\begin{equation}
	\mathcal{L}_{lip} = \lvert \nabla I_r^{lip} * M^{lip} \rvert \cdot exp(\text{-} \lvert \nabla I^{lip} \cdot M^{lip} \rvert)
	\label{equ.4}
\end{equation}
$\lvert \cdot \rvert$ and $\nabla$ means absolute value and differential operator. 

\noindent
\textit{\textbf{Reconstruction Loss.}}
We calculate a MSE loss between the result $I_r$ and the edited image $I^e$ as the reconstruction loss:
\begin{equation}
	\mathcal{L}_{rec} = \lvert \lvert I_r - I^e \lvert \lvert_2^2
	\label{equ.6}
\end{equation}

\noindent
\textit{\textbf{Perceptual and Style Loss.}} 
Perceptual loss \cite{Chen2017PhotographicIS}calculates the feature distance between the input image and the target image in a pre-trained VGG network \cite{Simonyan2014VeryDC}. Style loss \cite{Johnson2016PerceptualLF} adds a Gram matrix \cite{Gatys2015TextureSU} in perceptual loss to penalize differences in style like color, texture. The formula is as follows:
\begin{equation}
	\begin{aligned}
		&\mathcal{L}_{pcp} = \sum_l \lambda_l \Vert \Phi_l(I_f) - \Phi_l(I^e) \Vert_1  \\
		&\mathcal{L}_{style} = \sum_l \lambda_l \Vert gram(\Phi_l(I_f)) - gram(\Phi_l(I^e)) \Vert_1
	\end{aligned}
	\label{equ.7}
\end{equation}
$\Phi_l$ denotes the outputs of middle layers of a pretrained VGG network, and the weights $\lambda_l$ determine which layers of the network are used. $gram(\cdot)$ is a matrix of inner products of a set of vectors in an inner product space.

\noindent
\textit{\textbf{Adversarial Loss.}} 
We also use adversarial loss \cite{Mirza2014ConditionalGA} to increase image detail. The adversarial loss $\mathcal{L}_{adv}$ is calculated on the final result $I_f$ and the edited image $I^e$ through a learnable discriminator, which aims to distinguish the training data distribution and predicts one.

We first train the network using all losses other than lip-edge loss, the total loss is calculated by:
\begin{equation}
	\mathcal{L}_{total} = \lambda_{rec} \mathcal{L}_{rec} + \lambda_{pcp} \mathcal{L}_{pcp} + \lambda_{style} \mathcal{L}_{style} + \lambda_{adv} \mathcal{L}_{adv}
	\label{equ.8}
\end{equation}
After that, we finetune our method by adding lip-edge loss.

\begin{equation}
	\mathcal{L}_{ft} = \mathcal{L}_{total} + \lambda_{lip} \mathcal{L}_{lip} 
	\label{equ.9}
\end{equation}
We set $\lambda_{rec}=1,  \lambda_{pcp}=0.001, \lambda_{style}=10, \lambda_{adv}=0.01, \lambda_{lip}=0.1$.

\begin{table*}[t]
	\resizebox{1\linewidth}{!}
	{
		% Please add the following required packages to your document preamble:
		% \usepackage{booktabs}
		% \usepackage{multirow}
		\begin{tabular}{c|c|c|c|c|c|c|c|c|c}
			\hline
			Methods            & \multicolumn{1}{c|}{Type}                                                                        & \multicolumn{1}{c|}{\begin{tabular}[c]{@{}c@{}} $\uparrow$ Sync \\ Score\end{tabular}} & \multicolumn{1}{c|}{\begin{tabular}[c]{@{}c@{}} $\uparrow$ CLIP  \\ Direction\end{tabular}} & \multicolumn{1}{c|}{$\downarrow$ ArcFace} & \multicolumn{1}{c|}{$\downarrow$ Ew} & \multicolumn{1}{c|}{\begin{tabular}[c]{@{}c@{}}Temporal \\ Consistency\end{tabular}} & \multicolumn{1}{c|}{\begin{tabular}[c]{@{}c@{}}Pose \\ Manipulation\end{tabular}} & \multicolumn{1}{c|}{\begin{tabular}[c]{@{}c@{}}NVS / \\ Background\end{tabular}} & \begin{tabular}[c]{@{}l@{}}Details\\ Control\end{tabular} \\ \hline
			Wav2Lip            & \multirow{3}{*}{\begin{tabular}[c]{@{}c@{}}One-shot \\ Generation\end{tabular}}              &     \textbf{8.82}    &     0.0477   &   2.54    &   - & $\times$   &  $\times$ & $\times$ & $\times$ \\  
			MakeitTalk 
			&&    4.48    &     0.0518   &   2.36    &   0.0053 &  \checkmark   &  $\times$ & $\times$ & $\times$ \\  
			SadTalker    			                                                                                          &&    5.36    &     0.0442   &   2.86    &   \underline{0.0045} & \checkmark   &  $\times$ & $\times$ & $\times$ \\   \hline
			
			%			In2n & \multirow{4}{*}{\begin{tabular}[c]{@{}c@{}}Instruction-based \\ Neural Editing\end{tabular}} 
			%			&     -       &     0.0175        &   -      &   0.0048 & -      & \checkmark  &  \checkmark & \checkmark & $\times$ \\
			
			Ip2p  & \multirow{3}{*}{\begin{tabular}[c]{@{}c@{}}Instruction-based \\ Neural Editing\end{tabular}} 
			&    5.17    &     \underline{0.0597}   &   \textbf{1.81}    &   0.0120& $\times$    &  \checkmark & \checkmark & $\times$ \\    
			Ip2p + deflicker      
			&&    6.61    &     0.0435   &   2.16    &   0.0054  & $\times$    &  \checkmark & \checkmark & $\times$ \\
			\cline{1-1} \cline{3-10} 
			Ours             
			&&    \underline{6.66}     &     \textbf{0.0642}   &   \underline{1.99}    &   \textbf{0.0044} & \checkmark    &  \checkmark & \checkmark & \checkmark  \\ \hline
		\end{tabular}
	}
	\caption{Quantitative comparison results. Our method achieves top rankings in all four metrics. We show the best in bold and the second with underline. Moreover, our method supports various conditional controls such as pose manipulation, novel view synthesis, background replacement, and controllable details generation.}
	\label{comp_table}
\end{table*}

\begin{table}[t]
	\centering
	
	%	\resizebox{1\linewidth}{0.08\linewidth}
	\resizebox{1\linewidth}{!}
	{ 
		\begin{tabular}{l|c|c|c|c}
			\hline 
			
			\multirow{2}*{Methods} & Lip  & Instruction  & Face & Overall \\ 
			%			\cline{2-5}
			& Sync. &  Editing Quality &  Identity Preserving  & Video Quality \\
			%			Methods & Lip Sync. &  Instruction-Edited Quality & Face Identity Preserving & Overall Video Quality \\ 
			\hline
			
			Wav2Lip & 1.75\% & 1.75\% & 0.75\% &  1.25\% \\
			MakeitTalk  &  1\% & 2.75\% & 1.0\%  & 1.25\% \\
			SadTalker  & \underline{13.5\%}  & \underline{17\%}  & \underline{16.75\%}  & \underline{27\%}\\
			ip2p  & 5\% & 4.75\%& 6.25\% & 2.0\% \\
			ip2p+deflicker  & 8\% & 4.0\%& 8.5\% & 4.75\%\\
			\hline
			Ours  & \textbf{70.75\%} & \textbf{69.75\%} & \textbf{66.75\%} & \textbf{63.75\%} \\
			
			\hline
		\end{tabular}
	}
	\caption{User studies. Participants prefer our results best.}
	\label{user_study}
\end{table}

\section{Experiments}
\label{sec::results}
% 我们首先展示了一些基于指令编辑的结果，如Figure 中所示。通过输入不同的指令，Instruct-NeuralTalker可以编辑得到任意风格的Talking radiance field，从而渲染出高质量的说话人脸。Neural Radiance Field对跨视角一致性的要求使得我们的方法能够生成时序一致的输出。我们的训练策略和lip-edge loss在编辑后的神经辐射场中保持很好的音唇一致性。Please refer our supplementary video for better experience.
% 我们的方法支持使用简单的指令进行复杂的3D说话人脸编辑，极大地提升了对说话人脸生成的编辑能力并降低了编辑难度。收益于动态神经辐射场，我们能够基于音频控制神经辐射场生成高质量，时序一致，音唇同步的说话人脸。
\subsection{ Implementation Details and Metrics}

\textit{Datasets.}  
% 我们使用了一个经典的obama视频\cite{Suwajanakorn2017SynthesizingO}和来自于HDTF\cite{Zhang2021FlowguidedOT}的九个视频.每个视频长约5min。在初始化talking radiance field时，我们使用前90%的帧作为训练集进行训练，后10%的帧用于测试。在编辑过程中，我们只是用原来训练集中的少量图像进行训练。
We borrow the Obama video from \cite{Suwajanakorn2017SynthesizingO} and videos from HDTF \cite{Zhang2021FlowguidedOT} for our experiments. Each video has a duration of approximately 5 minutes. We take 1200 frames for initializing the talking radiance fields.

%For initializing the TRF, we use the first $90\%$ of frames from each video as the training set and reserve the remaining 10\% of frames for testing.

\textit{ Implementation Details.} Since our method introduces convolution-based refinement work, the random ray sampling strategy is unsuitable. Instead, we apply a patch-based ray sampling to facilitate spatial dependencies between rays. We randomly crop a $256\times256$ patch from the edited images as supervision. When fine-tuning our method with lig-edge loss, we only use a $64 \times 64$ patch of lip regions for training. Since the talking radiance field has already been initialized, we only need a few images to participate in the optimization during editing, further speeding up the model training. Typically, we set the number of images participating in training N to 200 and the dataset updating times $\mathbf{K} = 3$. We first train 300 epochs on a large patch and then finetune another 100 on a lip patch. We take Adam as the optimizer, and the learning rate is set as 5e-4 for MLP and 2e-4 for our refinement network, which takes roughly an hour on a single NVIDIA RTX 3090. More details are provided in the supplementary material. 

% 为了衡量编辑结果的音唇一致性，我们遵循之前的工作，使用the identity agnostic SyncNet confidence (Sync). 为了衡量编辑结果与文本目标是否一致，我们使用CLIP text image similarity来计算目标prompt和编辑结果的CLIP embedding的余弦距离。然而，由于我们的方法给定的是编辑指令，不能直接反应目标语义，所以我们手动给定编辑指令对应的前后prompt。例如编辑指令为“Make him look like a young man”, 我们设置编辑前的prompt为“A photograph of man”, 编辑后的指令为“A photograph of a young man”.
%同时我们还使用一个身份距离损失（ArcFace）来表示编辑后的人脸的身份保持。最后We measure the temporal inconsistency based on the a warping error \cite{Lei_2023_CVPR} that considers both short-term and long-term warping errors for quantitative evaluation. 
%We manually provide pre- and post-editing image captions corresponding to the editing instructions. For example, we provide "A photograph of a man" and "A photograph of a young man" for the instruction "Make him look like a young man".

%, which measures consistency of the change between the two images (in CLIP space) with the change between the two image captions.

\textit{ Evaluation Metrics.} We evaluate our method for editing equality on four main aspects. To assess the audio-lip synchronization, we follow previous works \cite{Guo2021ADNeRFAD} and employ the identity-agnostic SyncNet confidence (\textbf{Sync score} \cite{Chung_Zisserman_2017}). We evaluate the instruction editing quality with the directional similarity in CLIP space (\textbf{CLIP Direction} \cite{Gal2021StyleGANNADACD}). Additionally, we employ a face identity distance loss (\textbf{ArcFace} \cite{Deng2018ArcFaceAA}) to measure face identity preserving. For temporal inconsistency, we apply a warping error ($\textbf{E}_\textbf{w}$ \cite{Lei_2023_CVPR}) that considers both short-term and long-term warping errors. Please see the supplementary material for details.

\subsection{Comparisons}
\label{sec::results::comp}
\textit{Comparison settings.} Since there is no method to consider instruction-based talking face editing, we compare our method with two-stage methods, which consist of two main categories: (1) editing an image first using ip2p and then applying it to one-shot talking face generation methods (Wav2Lip\cite{10.1145/3394171.3413532}, MakeitTalk\cite{Zhou2020MakeItTalkST}, SadTalker\cite{zhang2022sadtalker}) (2) using RAD-NeRF to render the target image first, and then post-process it using ip2p. We also use the latest deflickering method\cite{Lei_2023_CVPR} to post-process the second type of method (ip2p+deflicker). 

\begin{table}[]
	\centering
	\resizebox{1\linewidth}{!}
	{
		\begin{tabular}{c|c|c|c|c}
			\hline
			Methods  & \begin{tabular}[c]{@{}l@{}} $\uparrow$ Sync \\ Score\end{tabular} & \begin{tabular}[c]{@{}c@{}}$\uparrow$ CLIP \\ Direction\end{tabular} & $\downarrow$ Ew                  &  \begin{tabular}[c]{@{}c@{}}Training \\ Time\end{tabular} \\ \hline
			Iterative dataset update   &        \textbf{6.73}                                               & 0.0175                                                    & \underline{0.0048}                                                                      & 5 h                                                      \\ \hline
			One time dataset update    &         4.36                                              &         \textbf{0.0699}                           &  0.0054                                                        & 1 h                                                       \\ \hline
			Progressive dataset update &         \underline{6.66}                                              & \underline{0.0642}                                                     & \textbf{0.0044}                                                                & 1 h                                                      \\ \hline
		\end{tabular}
	}
	\caption{Results and training time of three different dataset updating strategies.}
	\label{ablation_idu}
\end{table}

%
%\begin{table}[t]
%	\centering
%	%	\resizebox{1\linewidth}{0.08\linewidth}
%	\resizebox{1\linewidth}{!}
%	{ 
	%		\begin{tabular}{l|c|c|c|c}
		%			\hline 					
		%			%			\multirow{2}*{Metho   ds} & \multicolumn{4}{c|}{"Results on 10 "} \\ 
		%			%			\cline{2-5}
		%			Methods & $\uparrow$ Sync score & $\uparrow$ CLIP Direction & $\downarrow$ ArcFace & $\downarrow$ $E_w$ \\ 
		%			\hline
		%			Baseline & 3.604  & \textbf{0.1070} & 3.0580 & 0.0045 \\					
		%			Baseline+PDU & 4.597  & 0.0708 & 2.9642 &\textbf{ 0.0042} \\
		%			Baseline+PDU+$L_{lip}$ & \textbf{4.954} & 0.0442 & \textbf{2.6490} & 0.0063\\
		%			
		%			\hline
		%		\end{tabular}
	%	}
%	\caption{Ablation study comparing our full method with and without progressive dataset update (PDU) and lip-edge loss.}
%	\label{ablation_table}
%\end{table}

\begin{table}[t]
	\centering
	%	\resizebox{1\linewidth}{0.08\linewidth}
	\resizebox{1\linewidth}{!}
	{ 
		\begin{tabular}{l|c|c|c|c}
			\hline 					
			%			\multirow{2}*{Metho   ds} & \multicolumn{4}{c|}{"Results on 10 "} \\ 
			%			\cline{2-5}
			%			Methods & $\uparrow$ Sync score & $\uparrow$ CLIP Direction & $\downarrow$ ArcFace & $\downarrow$ $E_w$ \\ 
			Methods  & \begin{tabular}[c]{@{}l@{}} $\uparrow$ Sync \\ Score\end{tabular} & \begin{tabular}[c]{@{}c@{}}$\uparrow$ CLIP \\ Direction\end{tabular} & 
			$\downarrow$ ArcFace                  & 
			$\downarrow$ Ew                 \\ \hline
			
			\hline
			Base & 4.54  & 0.0430 & 2.62 & 0.0051 \\					
			Base + PDU & 6.33  & 0.0458 & 2.03 & 0.0043 \\
			Base + PDU + RefineNet($\omega$=0.4) & \underline{6.30} & \underline{0.0569}  & \underline{2.01}  & \textbf{0.0044}\\
			Base + PDU + RefineNet($\omega$=0.8) & \textbf{6.66} & \textbf{0.0642} & \textbf{1.99}  & \textbf{0.0044}\\
			
			\hline
		\end{tabular}
	}
	\caption{Ablation study for our method with and without progressive dataset update (PDU) and refinement network.}
	\label{ablation_table}
\end{table}

%Our method enables complex 3D talking face editing using simple instructions. Leveraging dynamic neural radiance fields, we can generate high-quality, temporally consistent, and audio-visual lip-synchronized talking faces based on the audio input.

\textit{Qualitative results.} We first illustrate some results of our method, as shown in Figure \ref{result}. Figure \ref{comparison} shows two editing results for comparisons. Wav2Lip can only produce a static head result. MakeitTalk generates a lip shape that differs significantly from the original result. SadTalker, the current SOTA one-shot-based talking face generation method, can produce high-quality results but does not match the original lip shape well. The post-processing-based network can restore a more consistent lip shape but suffers from flicker and distortion, while our method produces the best editing results. We highly recommend watching the video for a better visual experience.

%  wav2Lip专注于生成嘴型，因此它具有最高的音唇一致性。我们忽略评价它的视频一致性，因为它只生成静止的头部。MakeitTalke实现音唇一致性上表现很差。 SadTalker能够生成流畅的高质量的说话人脸视频，但是它并不能保持原来人脸的身份特性，也无法保证编辑目标。使用ip2p进行后处理的方法时序一致性很差，因为它产生闪烁的结果。经过去闪烁之后，生成质量有一定提升，但是也破坏了原始身份和编辑方向。
% 我们的方法所有四个指标上都有第一或者第二的表现。
\textit{Quantitative Results.} We show quantitative results in Table \ref{comp_table}. Wav2Lip only focuses on generating lip movements and achieving the highest audio-lip synchronization. We disregard evaluating its video consistency as it only generates static heads. MakeitTalk performs poorly in audio-lip consistency, and SadTalker can generate high-quality talking faces but fails to preserve the original identity and guarantee instruction editing direction. The method utilizing ip2p for post-processing exhibits poor temporal consistency. After deflickering, there is some improvement in generated quality, but it compromises the original identity and editing direction. Our method achieves top rankings in all four metrics, either first or second. In addition to instruction-based editing, our method supports a wide range of editing tasks, such as pose manipulation, novel view synthesis, background replacement, and controllable details generation. Visual results are shown in Figure \ref{detail generation}.

\textit{User Studies.} We conduct user studies to evaluate the quality of generated talking faces. We generate 20 videos from different characters and text instructions. We invite 20 participants and let them choose the best method for lip synchronization, instruction editing quality, face identity preserving, and overall video quality. We show the result in Table \ref{user_study}, where the participants prefer our method mostly in all four evaluation terms. We attribute this to the superiority of our method in both audio-lip consistency and video quality.
% 左侧展示了4种客观指标。在
%We show the quantitative results in Table \ref{comp_table}. 我们对不同人物进行了共10种编辑，并在表的左侧汇报4种客观指标。在音唇一致性上，我们的方法除了逊色于专门生成唇形的wav2lip，比其他方法都要好。

%
%\begin{table}[t]
%	\centering
%	\caption{.}
%	\resizebox{1\linewidth}{0.08\linewidth}
%	{ 
	%		\begin{tabular}{l|c|c|c|c}
		%			\hline 
		%			
		%			\multirow{2}*{Methods} & \multicolumn{4}{c|}{"Turn him into a statue"} \\ 
		%			\cline{2-9}
		%			& $\uparrow$ Sync score & $\uparrow$ CLIP Direction & $\downarrow$ ArcFace & $\downarrow$ $E\_w$ \\ 
		%			\hline
		%			
		%			Ours  & 4.937 & \textbf{-0.0008} & \textbf{2.6233} & 0.0063 \\									
		%			
		%			\hline
		%		\end{tabular}
	%	}
%	%	\caption{.}
%	\label{ablation}
%\end{table}

%
%\begin{figure}[t]
%	\centering
%	\includegraphics[width=0.9\linewidth]{imgs/nvs_bg.pdf}
%	\caption{.}
%	\label{nvs_bg}
%\end{figure}

\begin{figure}[t]
	\centering
	\includegraphics[width=\linewidth]{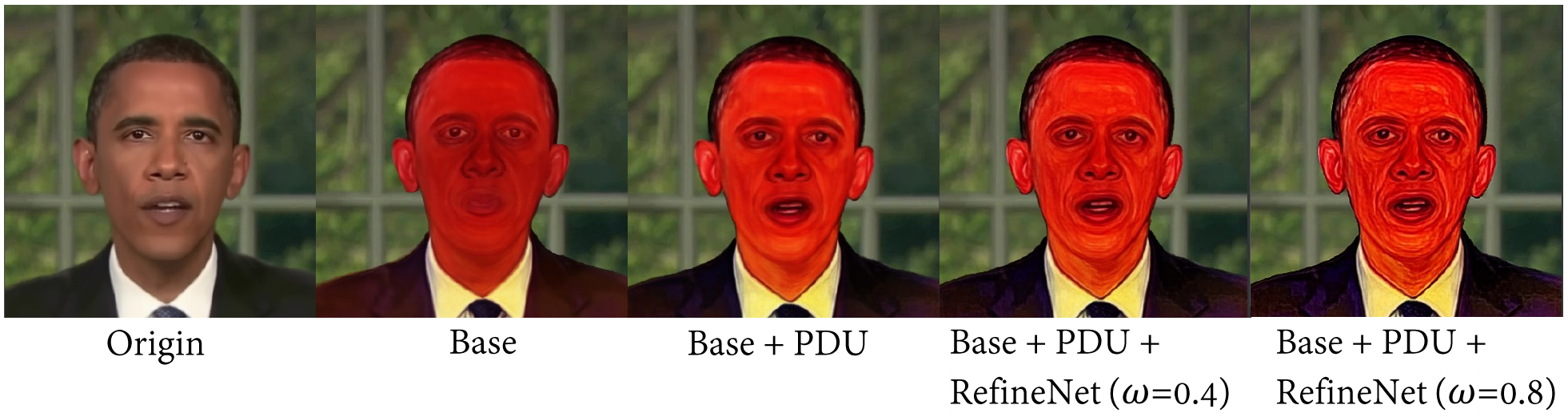}
	\caption{Illustration of ablation study with instruction \textit{"Make him look like an Edward Munch painting"}.}
	\label{ablation}
\end{figure}

\subsection{Ablation study}
We first validate our method by comparing it to different dataset updating strategies. Figure \ref{pdu} and Table \ref{ablation_idu} show the qualitative and quantitative results. Iterative dataset update is very slow and not able to guarantee successful editing, shown in the little difference between the original image and low \textbf{CLIP Direction} value. Although one time dataset update speeds up training time, we find that it destroys the lip shape, shown in the low \textbf{Sync Score} value. Our progressive dataset update achieves successful editing while maintaining high audio-lip synchronization.

%Iterative dataset update proposed by Instruct-NeRF2NeRF edit one image in the dataset at a time. It is very slow and not able to guarantee successful editing, which result in little difference with the original image. We attribute it to the extra image conditions in diffusion model and the sparse views in the talking face. We remove the extra image condition and perform one time dataset update, which edit the whole dataset once in the beginning. Although it speeds up training time and allows for successful editing, we find that it may destroy the lip shape leading to low audio-lip synchronization. Our progressive dataset update can

%Iterative dataset update is very slow and not able to guarantee successful editing, which result in little difference with the original image. Although one time dataset update speeds up training time and allows for successful editing, we find that it may destroy the lip shape leading to low audio-lip synchronization.
We then evaluate each component in our method. We take one time dataset update as a baseline and add the progressive dataset update strategy (PDU) and refinement network to it step by step. We show the quantitative results in Table \ref{ablation_table}. PDU brings a significant improvement in audio-lip synchronization. We also find that increasing the refinement weight is helpful to a higher CLIP Direction value as it enriches the image details. Figure \ref{ablation} show some visual results. Our baseline method destroys the lip's shape. By adding PDU, the lip structure matches the original image well. By adding refinement networks and controlling the weights, we can adjust the details of the final results. We also conduct experiments on the efficiency of rendering. The rendering speed is approximately 30 FPS when the refinement network is off and 25 FPS when the refinement network is enabled.

\section{Conclusion}
\label{sec::Conclusion}

%\textit{Limitation.} 
%% 我们方法的编辑能力主要受限于ip2p基于指令的编辑能力。由于ip2p没有在人脸数据上进行优化，因此它对人脸的编辑能力有限，会导致一些错误的编辑。因此在未来探索对人脸图像的指令编辑会是一个有趣的方向。此外，我们的方法必须经过优化才能执行编辑，无法快速推广到任意的指令。探索无需优化的3D人脸编辑方法是我们下一步的目标。
%Our method's editing capability is primarily limited by ip2p. As ip2p has not been specifically optimized for facial data, it has limited ability to edit faces, which has no face structure prior and may leads to some wrong edits. Therefore, exploring instruction-based editing on facial images would be an interesting direction for future research. Additionally, our method requires optimization to perform edits and cannot be rapidly extended to arbitrary instructions. Exploring optimization-free 3D facial editing methods is our future network.
% 我们提出了Instruct-NeuralTalker，第一个新颖的互动框架，允许使用指令编辑Talking radiance field来生成个性化的说话人脸。Instruct-NeuralTalker极大地拓展了对3D Talking face的编辑能力，能够实现出了视角合成和背景替换之外的更加广泛的编辑. 为了保持编辑结果的音唇同步性，我们引入了progressive dataset update stragety and lip-edge loss来约束嘴部形状。我们还引入了一个refinement network来克服过度平滑的结果，并支持实现可控的细节生成。我们的方法支持在消费级硬件上实现实时渲染。最后，多种评价指标证明了我们方法相比于SOTA能够实现更加高质量的编辑。
% to generate personalized talking faces. 
%\noindent
%\textit{Conclusion.} 
We present Instruct-NeuralTalker, the first interactive framework to edit talking radiance fields with instructions. Instruct-NeuralTalker greatly expands the ability to edit 3D talking faces. It enables users to generate personalized talking faces with their instructions. In order to keep the audio-lip synchronization, we introduce a progressive dataset update strategy to keep the lip shape. We also introduce a refinement network to overcome over-smoothed results and support controllable detail generation. In addition, our approach achieves real-time rendering on consumer hardware. 

%Finally, multiple evaluation metrics and user studies demonstrate that our method achieves higher quality editing than the state-of-the-art.

\bibliography{aaai24}

\clearpage

\begin{appendix}

\section{More Visualization}
\begin{figure*}[t]
	\centering
	\includegraphics[width=\linewidth]{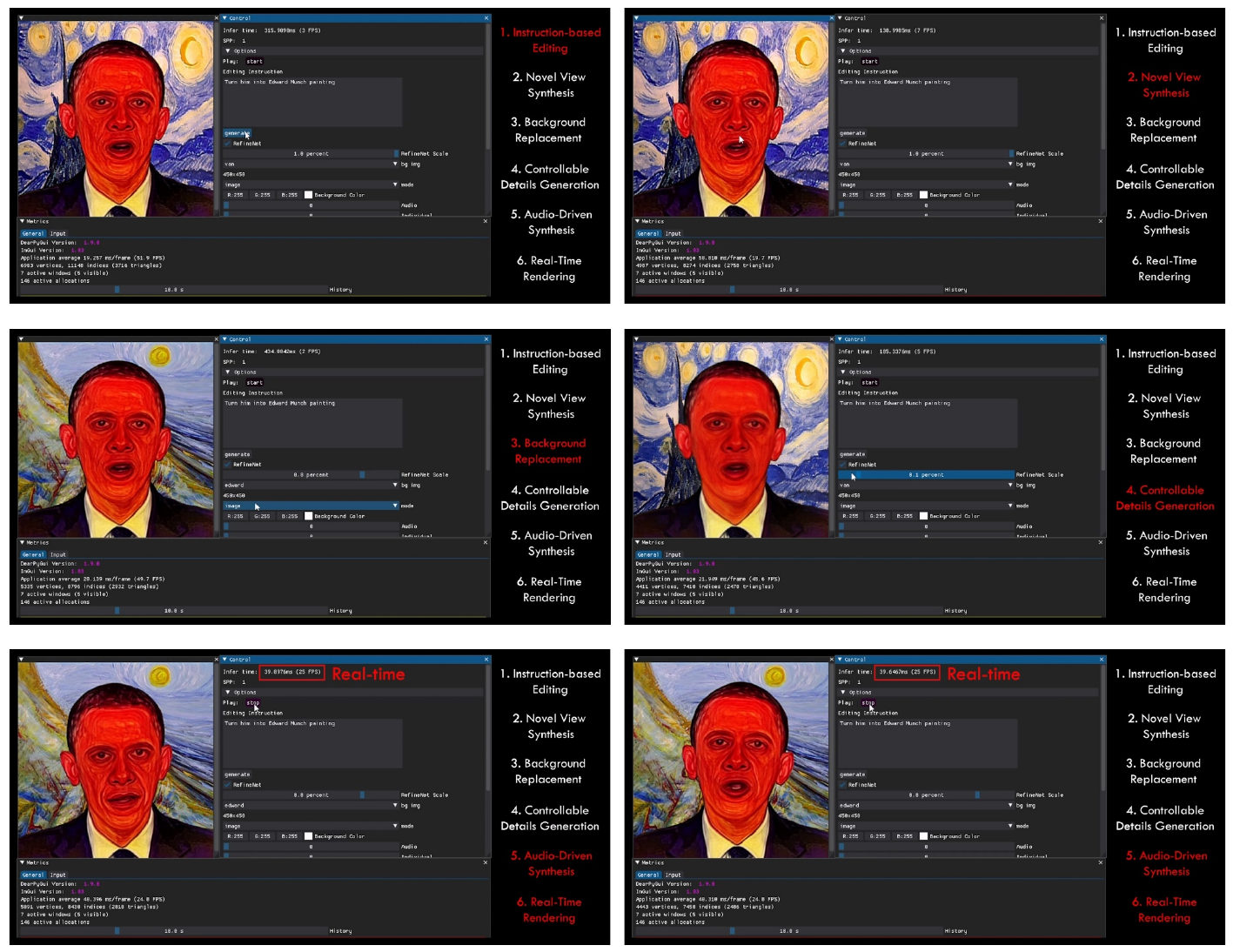}
	\caption{User interface. We implement an interactive interface for user to perform instruction-base editing. }
	\label{interface}
\end{figure*}

We implement an interactive interface for user to perform instruction-based editing, shown in Figure \ref{interface}. Our method achieves six functions : instruction-based editing, novel view synthesis, background replacement, controllable detail generation, audio-driven talking face synthesis and real-time rendering. More visual results are represented in Figure \ref{result}. For a better visual experience, we highly recommend watching the video demo in the supplementary files.

\section{Refinement Network}
We draw inspiration from the residual dense block (RRDB) in the super-resolution method ESRGAN \cite{Wang2018ESRGANES} to implement our refinement network. A single RRDB block consists of five convolutional layers, each following a leaky relu activation function. A skip connection exists between every two convolutional layers. To keep the implementation lightweight, we use only one RRDB module, with a model size of only 1MB, which allows us to achieve real-time rendering on consumer-grade hardware. We sum the results of the refinement network and the direct rendering of the talking radiance field according to the weights $\omega$ so that it allows for controlled detail generation. We set the $\omega = 0.8$ during the optimization.

\section{Implementation Details and Metrics}
\textit{Data pre-processing.} We borrowed the Obama video from \cite{Suwajanakorn2017SynthesizingO} and nine other videos from HDTF \cite{Zhang2021FlowguidedOT} for the experiment. The duration of each video is 5 minutes. We first extract face landmarks and parse masks by landmark detection and face parsing. Then, we use the landmarks to optimize the 3D MM model to obtain the head pose, expression vectors, and so on. For audio processing, an Automatic Speech Recognition (ASR) model \cite{Baevski2020wav2vec2A} is applied to extract audio features from the audio track.

\textit{Evaluation Metrics.} We evaluate our method for editing equality on four main aspects. We take a closer look at CLIP Direction and $E_w$ in the following:

$\bullet$ CLIP Direction. We evaluate the instruction editing quality with the directional similarity in CLIP space (CLIP Direction \cite{Gal2021StyleGANNADACD}). We manually provide pre- and post-editing image captions corresponding to the editing instructions. For example, we provide "A photograph of a man" and "A photograph of a young man" for the instruction "Make him look like a young man".

\begin{figure}[t]
	\centering
	\includegraphics[width=0.9\linewidth]{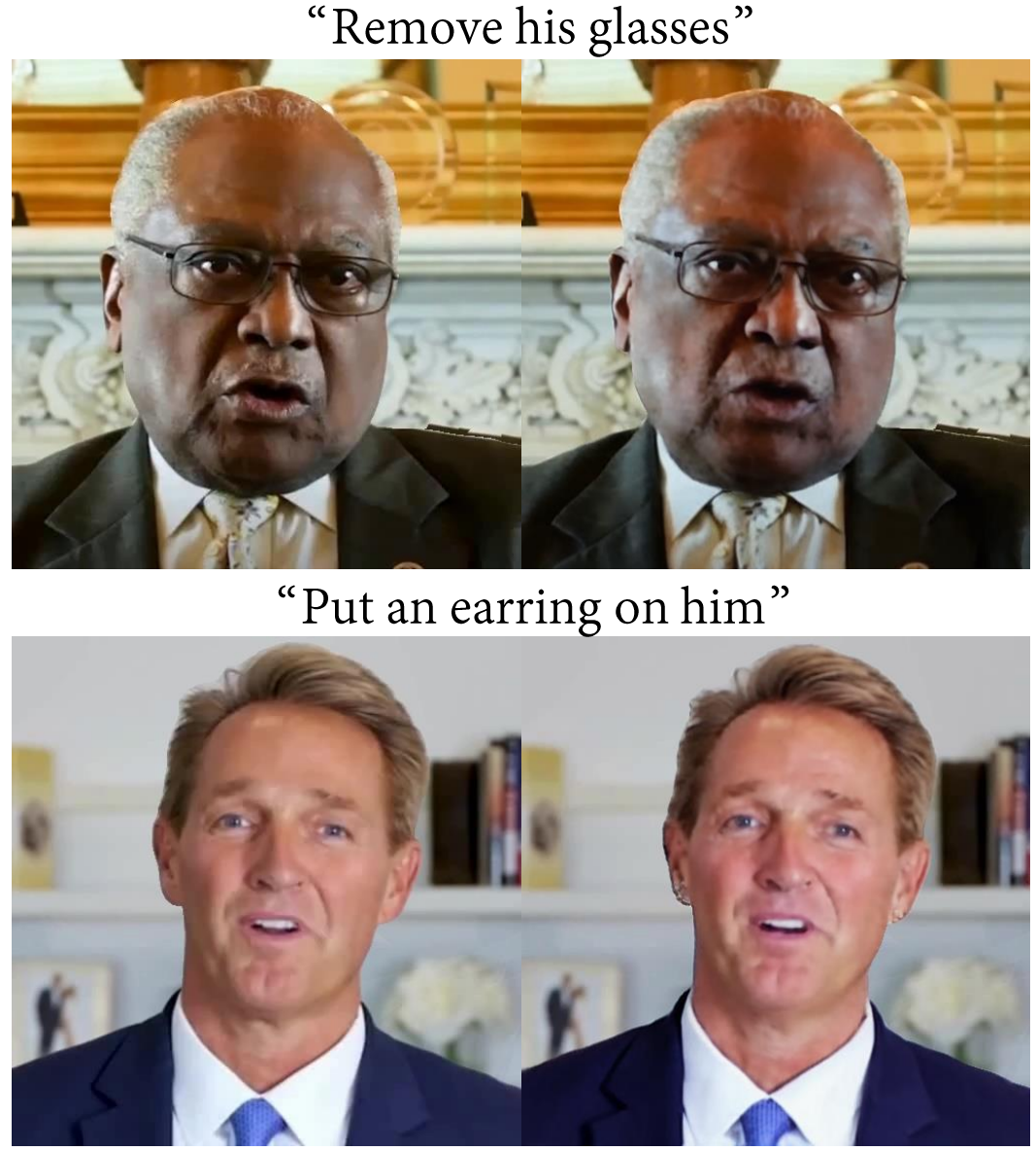}
	\caption{Bad cases of our method. The left column is the original image, and the right colume is the edited result by ip2p. Ip2p struggles with adding and removing objects. Fine-tuning ip2p in a instruction-based face editing dataset may sovle this problem.}
	\label{badcase}
\end{figure}

\begin{figure*}[t]
	\centering
	\includegraphics[width=\linewidth]{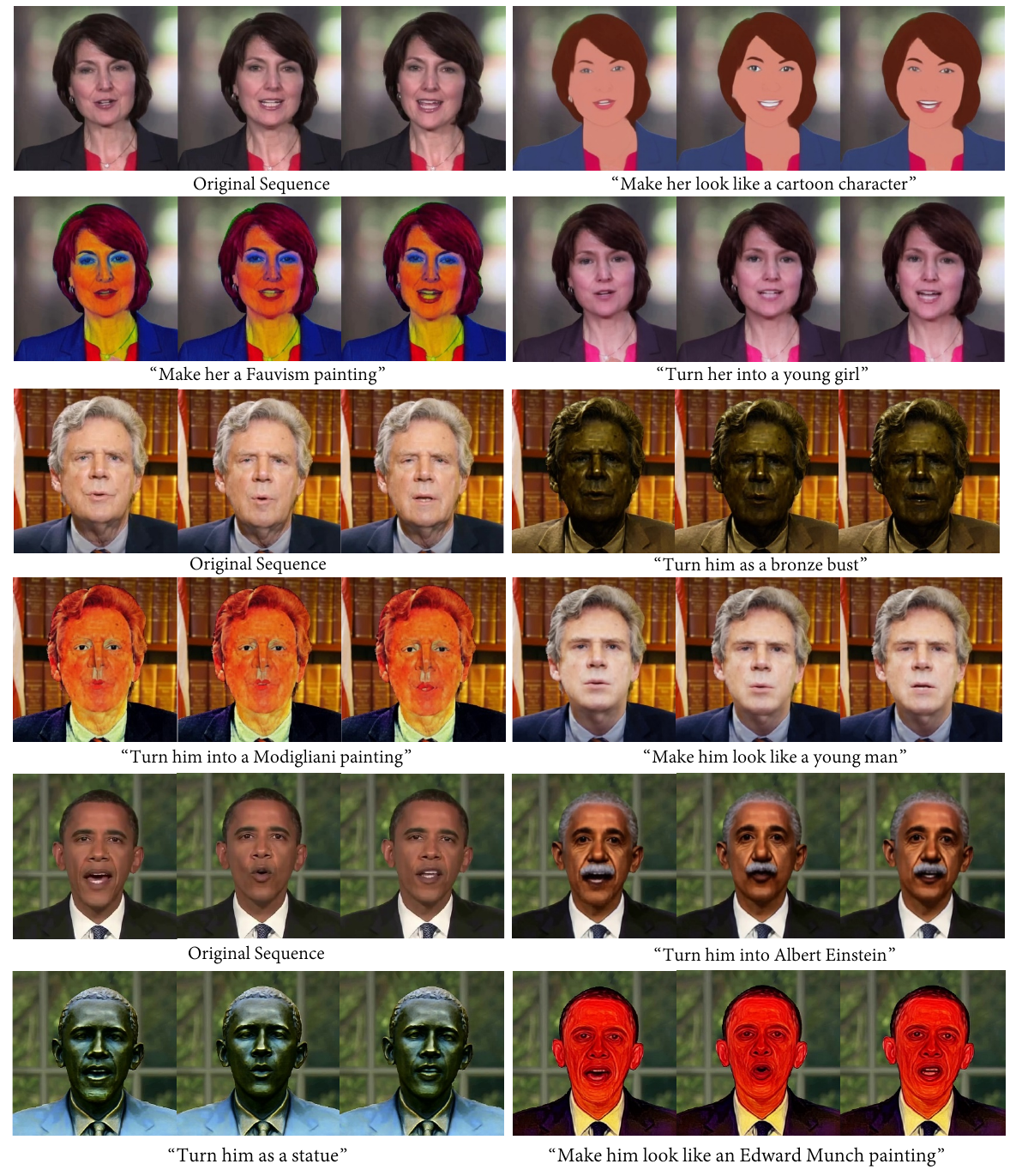}
	\caption{Visual results of our method.}
	\label{result}
\end{figure*}

$\bullet$ $E_w$. For temporal inconsistency, we apply a warping error ($E_w$ \cite{Lei_2023_CVPR}) that considers both short-term and long-term warping errors. Given a pair of frames $O_t$ and $O_s$, the warping error $E_{pair}$ can be calculated by:
\begin{equation}
	\begin{aligned}
		& E_{pair}(O_t, O_s) = \lvert \lvert M_{t,s} \odot (O_t - \hat{O_s}) \lvert \lvert_1 \\
		& E_{warp}^t = E_{pair}(O_t, O_{t-1}) + E_{pair}(O_t,O_1)
	\end{aligned}
	\label{equ.7}
\end{equation}
where $\hat{O_s}$ is obtained by warping the $O_s$ with the optical flow from frame $t$ to frame $s$. $M_{t,s}$ is the occlusion mask from frame $t$ and frame $s$. For each frame $t$, the warping error $E_{warp}^t$ is computed with the previous and first frames in the video. We use PWC-Net \cite{Sun2017PWCNetCF} to calculate the optical flow between two frames.

\textit{Implementation Details.}
Our progressive dataset update strategy fix the image gudiance scale $s_I=1.5$, and determine text guidance scale $s$ and diffusion step by 
uniformly sampling from a lower and upper bound $[s_l, s_u]$, $[t_l, t_u]$. For most of the editing results in the paper, we set the $s_l = 3.0, s_u = 6.0$, $t_l = 10, t_u = 20$ and the dataset updating time $K=3$. For some other cases, we manually make adjustments to ensure the image quality of the editing result. In practice, the optimal choice for this bound is a subjective decision — a user may prefer more subtle or more extreme edits that are best found at different stages of training.

\section{Limitations.}
Our method inherits many of the limitations of InstructPix2Pix. For example, it struggles with counting numbers of objects and with spatial reasoning and does not support adding or removing large objects. We show some base cases in Figure \ref{badcase}. Our method's editing capability is primarily limited by ip2p. As ip2p has not been specifically optimized for facial data, it has limited ability to edit faces, which has no face structure prior and may leads to some wrong edits. Therefore, exploring instruction-based editing on facial images would be an interesting direction for future research. Additionally, our method requires optimization to perform edits and cannot be rapidly extended to arbitrary instructions. Exploring optimization-free 3D facial editing methods is our future network. 

\end{appendix}

\end{document}